# LTPNet Integration of Deep Learning and Environmental Decision Support Systems for Renewable Energy Demand Forecasting

Te Li[1,2*], Mengze Zhang[3*], Yan Zhou[4]

[1]School of Business Administration, Ningbo University of Finance & Economics, Ningbo 315175, China

[2]School of Business Administration, Dongbei University of Finance & Economics, Dalian 116025, China

[3]School of Digital Technology and Engineering, Ningbo University of Finance & Economics, Ningbo 315175, China

[4]Northeastern University San Jose, CA, 95131, USA

*Corresponding author Email: richte82@163.com; zmzyanzi@163.com

## ABSTRACT

Against the backdrop of increasingly severe global environmental changes, accurately predicting and meeting renewable energy demands has become a key challenge for sustainable business development. Traditional energy demand forecasting methods often struggle with complex data processing and low prediction accuracy. To address these issues, this paper introduces a novel approach that combines deep learning techniques with environmental decision support systems. The model integrates advanced deep learning techniques, including LSTM and Transformer, and PSO algorithm for parameter optimization, significantly enhancing predictive performance and practical applicability. Results show that our model achieves substantial improvements across various metrics, including a 30% reduction in MAE, a 20% decrease in MAPE, a 25% drop in RMSE, and a 35% decline in MSE. These results validate the model's effectiveness and reliability in renewable energy demand forecasting. This research provides valuable insights for applying deep learning in environmental decision support systems.

## 1 Introduction

In the context of the rapid growth of global energy demand and the huge challenges brought by climate change, the green operations and sustainable development of enterprises have become the common goals of all industries. Accurately forecasting and meeting renewable energy needs is not only an important way for enterprises to achieve green operations, but also the key to addressing environmental changes and fulfilling their social responsibilities. However, traditional energy demand prediction methods usually show limited capacity in processing complex data, and the prediction accuracy is low, which cannot meet the precise management needs of enterprises under dynamic market conditions. This situation makes enterprises face huge uncertainties and risks when formulating energy management strategies(Shekhar et al., 2023).

With the advancement of the global sustainable development agenda and the importance of countries on environmental protection and emission reduction, how to more effectively use renewable energy to ensure the stability and reliability of energy supply has become a practical problem for enterprises to solve(Fan et al., 2023). In this context, the development of new forecasting methods to improve the accuracy and reliability of energy demand forecasting is the key to promoting the green business operations and promoting the realization of global emission reduction targets.

In recent years, deep learning technology has provided new solutions for energy demand prediction due to its powerful data processing power and excellent prediction performance. Deep learning models show great potential in capturing complex energy demand patterns, and can provide more reliable decision support for the sustainable operation of enterprisesd(Ying et al., 2023). However, the application of existing deep learning models in environmental decision support systems faces multiple challenges, especially with limitations in capturing long-term dependencies in time-series data. Therefore, an approach that integrates deep learning technologies with environmental decision support systems is needed to better predict future renewable energy needs and support enterprises to develop more scientific energy management strategies(Wang et al., 2024).

Based on the above questions, this study presents an innovative approach combining LSTM, Transformer model and PSO. This method aims to overcome the limitations of traditional prediction models and improve the accuracy and reliability of renewable energy demand prediction through the integration of advanced deep learning techniques and optimization algorithms.

The primary contributions of this research are as follows:

First, this paper introduces a method that integrates deep learning technology with environmental decision support systems to predict renewable energy demand for enterprises. Incorporating advanced deep learning techniques, including LSTM and Transformer models, allows the system to more effectively capture temporal dependencies and long-term trends in energy demand, thereby improving the accuracy and reliability of predictions.

Secondly, it utilizes the PSO algorithm to optimize model parameters, thereby further enhancing predictive performance. The PSO algorithm efficiently explores the parameter space to identify the optimal combination, rendering the overall model more adaptable to practical applications and improving its generalization capabilities.

Finally, it constructs a comprehensive forecasting and optimization framework that provides reliable decision support for green operations. This framework not only accurately forecasts future renewable energy demand but also assists companies in crafting more sustainable energy management strategies. It aids in reducing energy consumption, diminishing reliance on traditional energy sources, and thereby advancing environmental protection and sustainable development.

Following this introduction, the paper proceeds as follows: Section 2 provides a comprehensive literature review on this topic from both domestic and international perspectives. Section 3 elaborates on the methods and model design used in this study, including the implementation details of LSTM, Transformer model, and the PSO algorithm. Section 4 introduces the experimental setup and analyzes the results to validate the effectiveness and performance of the proposed model. The final section provides a summary of the paper and discusses future research directions and potential developments in this field.

## 2 Related Work

In the field of smart city traffic carbon emission prediction, the application of deep learning model has made significant progress. This section will introduce the research progress and achievements of different deep learning models in smart city traffic carbon emission prediction in four parts.

### 2.1 Renewable Energy Demand Prediction Methods

In green operations, an accurate forecast of renewable energy demand is a crucial step toward sustainable development. Over the past few years, numerous methods have emerged to forecast the renewable energy demand of enterprises(Arumugham et al., 2023). These methods aim to utilize historical data and various models to predict future energy demand, thereby assisting companies in formulating more effective energy management strategies(Wang et al., 2024).

Current research primarily aims to improve prediction accuracy and adaptability. In addition to traditional deep learning models, several statistical and machine learning-based approaches are widely used(Al-Janabi & Al-Janabi, 2023). For example, models like Support Vector Machines (SVM), Multilayer Perceptrons (MLP), Adaptive Neuro-Fuzzy Inference

Systems (ANFIS), and Long Short-Term Memory (LSTM) networks have shown significant promise in energy demand forecasting(Abbasimehr et al., 2023; Qin et al., 2023). These methods, through modeling and analyzing data features, can more accurately capture the trends and patterns of energy demand fluctuations(Chaudhary et al., 2022; Li & Xiao, 2023).

These methods have significant advantages, such as more accurate prediction results and generalization capabilities. However, they also face several limitations, particularly when dealing with varying factors and complex environmental conditions. Traditional models often struggle to fully capture the nonlinear relationships and temporal dependencies present in the data, resulting in instability and volatility in prediction results. Therefore, more comprehensive and effective approaches are needed to address these issues and achieve accurate predictions of enterprises' renewable energy demand.

**2.2 Deep Learning in Renewable Energy Forecasting**

Recent years have seen increasing utilization of deep learning technology in renewable energy forecasting. With its powerful data processing capabilities and predictive performance, deep learning has shown tremendous potential in energy demand forecasting(Li et al., 2023). Within the context of green operations, deep learning technology can accurately capture the complex patterns of energy demand fluctuations, offering reliable decision support for enterprises.

Current research primarily aims to enhance enhancing deep learning models' predictive performance and adaptability. Commonly used models include Convolutional Neural Networks (CNN), Gated Recurrent Units (GRU), and Transformers, all of which have demonstrated notable success in energy demand forecasting(Mubarak et al., 2023). For instance, CNNs excel in extracting features from spatio-temporal data, while RNNs and their variants like LSTM and GRU specialize in capturing long-term dependencies in time series data(Joshi et al., 2022).

These models have significant advantages, such as more accurate prediction results and generalization capabilities. However, they also face several limitations, particularly in dealing with nonlinear relationships and temporal dependencies(Liu et al., 2023). Traditional deep learning models frequently encounter difficulties in effectively capturing the intricate relationships present within the data, which consequently results in instability and volatility in prediction outcomes(H. Zhang et al., 2024). Therefore, further improvements in deep learning models are needed to enhance their ability to process time series data, thereby better predicting the renewable energy demand of enterprises (Yao & Liu, 2024).

**2.3 Integration of Decision Support Systems**

Emerging research focuses on the integrated application of decision support systems (DSS) in environmental management. This integration seeks to merge deep learning technology with DSS, to boost decision-making efficiency and to elevate energy management standards within green operations(Zhou & Lund, 2023).

Currently, research focuses on optimizing the performance and adaptability of DSS. Typical integrated models include rule-based systems, knowledge-based systems, and data-based systems. These models have been widely applied in environmental management(Sahoo & Goswami, 2023). For example, rule-based systems can classify and analyze environmental data according to predefined rules, knowledge-based systems leverage expert knowledge bases to solve complex environmental issues, and data-based systems utilize big data analysis for environmental decision-making(Ning et al., 2024).

In recent years, the application of deep learning in environmental decision support systems (EDSS) has made remarkable progress. New research shows that models such as Transformer and the GNN perform well in processing data from complex environments and improving predictive accuracy(J.-L. Zhang et al., 2024) . In addition, hybrid models that combine deep learning and optimization algorithms also show great potential in environmental prediction. For example, genetic algorithms, particle swarm optimization (PSO), and Bayesian optimization methods have been successfully used to optimize the hyperparameters of deep neural networks and reinforcement learning models, improving the efficiency of air quality prediction and smart grid scheduling(Erden, 2023; Hu et al., 2022).

In the field of renewable energy prediction, the development of multimodal data fusion and time series modeling technology has also attracted wide attention. Research shows that by integrating CNN, LSTM and its variants, it can more effectively capture complex patterns of energy demand and significantly improve prediction performance(Chu et al., 2020; Chung et al., 2020; Sun et al., 2022). In addition, the model based on the Attention mechanism shows excellent performance in multi-source data fusion and multi-scale analysis, and is suitable for energy management decisions in complex environments(Bou-Rabee et al., 2022).

This integration has significant advantages, such as improved decision-making efficiency and accuracy. Nevertheless, they encounter significant limitations, particularly when dealing with vast datasets and complex environmental circumstances(Shahzad et al., 2023). Traditional DSS often struggle to fully utilize deep learning technology to address complex issues in environmental management, resulting in lower decision-making efficiency and accuracy (Wang, Sun, & Guo, 2024). Therefore, further research and development of integrated DSS are needed to better address the challenges enterprises face in green operations.

## 3 Methodology

### 3.1 Overview of LRF-PSO Model

The LTPNet model is the overarching framework for the renewable energy demand forecasting model proposed in this paper. LTPNet The model integrates three modules, LSTM, Transformer and PSO. Each of them is designed with its specific purpose and advantages: long-short-term memory network (LSTM) is designed to process time series data, which can effectively capture the long-term and short-term dependencies in historical data. LSTM was selected because it is able to make full use of the time-dependent characteristics in the data to improve the forecasting accuracy in dealing with complex time-series forecasting problems (such as renewable energy demand forecasting). Transformer The feature representation extracted by LSTM is further optimized by capturing global features through the self-attention mechanism (self-attention). Transformer Has the ability to effectively process long sequence data without relying on sequence calculation, which not only improves the computing efficiency, but also better captures the complex relationships in the input data, thus enhancing the robustness and generalization ability of the model. The Particle swarm optimization (PSO) algorithm is used to optimize the hyperparameters of the LSTM and Transformer modules to ensure that the model achieves the best performance. With Its powerful global search ability and good convergence, it can explore the parameter space efficiently without explicit gradient information, find the optimal hyperparameter combination, and then improve the prediction performance and practical application value of the model.

The LSTM component, which serves as the foundation part of the LTPNet model, processes historical energy consumption data. Through its unique structure, the LSTM model transforms historical data into feature representations to effectively capture temporal dependencies and long-term trends crucial for accurate predictions. Secondly, the Transformer component takes the features extracted by LSTM as input and utilizes its self-attention mechanism to further optimize the feature representations. The Transformer model demonstrates robust modeling capabilities by dynamically capturing relationships among various features, thereby boosting prediction accuracy and resilience. Finally, the PSO algorithm acts as the optimization module, adjusting the hyperparameters of both LSTM and Transformer to further improve the overall predictive performance. PSO efficiently explores the parameter space to identify the most optimal parameter combination, thereby enhancing the model's suitability for real-world applications.

The design of the LTPNet model encompasses the following essential stages. First, the model is structured with two layers of LSTM units to process historical energy consumption data from the past 24 months. This setup effectively captures both the temporal dependencies and long-term trends, thereby extracting valuable features for accurate predictions. Next, these features are fed into the Transformer component for further optimization and feature representation. The Transformer module employs a six-layer encoder-decoder structure, with each layer having eight attention heads. Residual connections and layer normalization are applied to stabilize the training process and speed up model convergence. Subsequently, the output of the Transformer is processed by a two-layer feedforward neural network, further extracting features and making the final energy demand predictions.In conclusion, the PSO algorithm optimizes the hyperparameters of both the LSTM and Transformer models to enhance overall predictive performance. PSO is utilized to search for the optimal combination of hyperparameters, significantly improving the model's forecasting accuracy.Figure 1 illustrates the entire process.

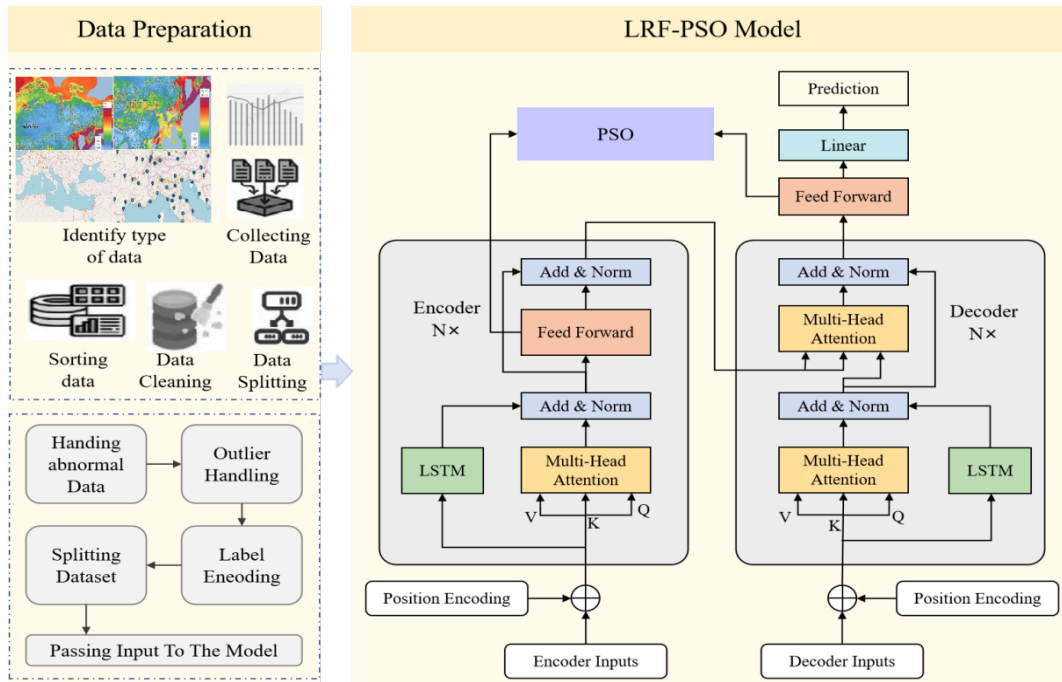

**Figure 1.Overall framework and data preparation process of LRF-PSO model.**

This figure shows the overall framework of LRF-PSO model, including the data preparation stage and the main structure of the model. In the data preparation stage, different types of data (such as wind power, solar energy and market transaction data) are first identified and collected, and then data sorting, cleaning, exception handling and data segmentation are performed to ensure data integrity and consistency. The processed data is input into the model. In the model structure part, the LRF-PSO model consists of multi-layer LSTM and Transformer encoder-decoder components, and uses a multi-head attention mechanism to capture the complex dependencies between data. LSTM is used to process the long-term and short-term dependencies of time series data, and the Transformer part further optimizes the feature representation. The PSO (particle swarm optimization) algorithm is used to adjust the hyperparameters of the LSTM and Transformer components to improve the prediction performance. The final model is predicted and output through a linear layer and a feed forward neural network.

The LTPNet model integrates LSTM, Transformer, and PSO to create a comprehensive framework for forecasting and optimization. Each component plays a unique role in the prediction process, collaborating synergistically to ensure precise forecasts of renewable energy demand for enterprises. Through the organic combination of these three components, the LTPNet model establishes a robust framework for forecasting and optimization, providing reliable decision support for green operations in enterprises.

**3.2 LSTM Component**

Long Short-Term Memory (LSTM) networks are specialized architectures designed to address the vanishing gradient problem commonly encountered in traditional Recurrent Neural Networks (RNNs). LSTMs effectively capture and retain long-term dependencies within sequential data, making them highly effective for modelling time series data, such as historical energy consumption record(Jailani et al., 2023). In the LTPNet model, the LSTM component is primarily responsible for processing historical energy consumption data and extracting pertinent features to capture temporal patterns and dependencies within the dataset.

LSTMs are widely utilized in renewable energy demand forecasting due to their efficacy in modeling and predicting intricate temporal dynamics. Their recurrent nature enables them to learn from past observations and capture intricate patterns in energy consumption behavior over time(Balakumar et al., 2023). Compared to traditional time series forecasting methods, LSTMs offer several advantages, including their ability to handle nonlinear relationships, adapt to different sequence lengths, and learn long-term dependencies.

In this study, the LSTM component plays a crucial role in processing historical energy consumption data, which forms the foundation for renewable energy demand forecasting. Based on analysis of past consumption patterns and trends, the LSTM extracts meaningful features that encode important information about energy usage behavior. These features are then passed to subsequent components of the LTPNet model for further processing and prediction. The LSTM component's capability to apprehend temporal dependencies and long-term trends in energy consumption data is crucial for precise forecasting of renewable energy demand and bolstering sustainable energy management practices in green operations. Additionally, Figure 2 illustrates the flowchart of the LSTM model, detailing its internal operational mechanisms.

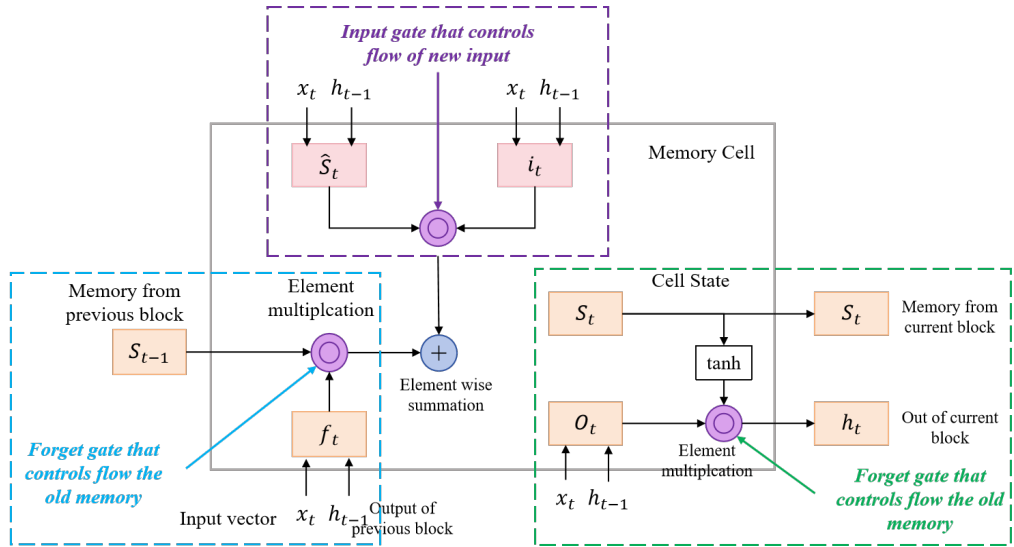

**Figure 2.** Flowchart of the LSTM model. The LSTM model controls the flow of information and the updating of memory through three main gate units (input gate, forget gate, and output gate). The input gate determines the amount of new input information; the forget gate adjusts the degree of retention or forgetting of old memory information; and the output gate controls the amount of information output by the current memory unit.

The main formula of LSTM model is as follows:

$$i_t = \sigma(W_{xi}x_t + W_{hi}h_{t-1} + W_{ci}c_{t-1} + b_i) \quad (1)$$

Where $i_t$ is the input gate activation vector at time step $t$. $x_t$ is the input vector at time step $t$. $h_{t-1}$ is the hidden state vector at time step $t-1$. $c_{t-1}$ is the cell state vector at time step $t-1$. $W_{xi}$, $W_{hi}$ and $W_{ci}$ are the weight matrices for input, hidden state, and cell state respectively. $b_i$ is the bias vector for input gate.

$$f_t = \sigma(W_{xf}x_t + W_{hf}h_{t-1} + W_{cf}c_{t-1} + b_f) \quad (2)$$

Where $f_t$ is the forget gate activation vector at time step $t$. $W_{xf}$, $W_{hf}$ and $W_{cf}$ are the weight matrices for input, hidden state, and cell state respectively. $b_f$ is the bias vector for forget gate.

$$c_t = f_t \odot c_{t-1} + i_t \odot tanh(W_{xc}x_t + W_{hc}h_{t-1} + b_c) \quad (3)$$

Where $c_t$ is the cell state vector at time step $t$. $\odot$ is the element-wise multiplication operator. $W_{xc}$ and $W_{hc}$ present the weight matrices for input and hidden state respectively. $b_c$ is the bias vector for cell state.

$$o_t = \sigma(W_{xo}x_t + W_{ho}h_{t-1} + W_{co}c_{t-1} + b_o) \tag{4}$$

Where $o_t$ is the output gate activation vector at time step $t$. $W_{xo}$, $W_{ho}$ and $W_{co}$ present the weight matrices for input, hidden state, and cell state respectively. $b_o$ is the bias vector for output gate.

$$h_t = o_i \odot tanh(c_t) \tag{5}$$

Where $h_t$ is the hidden state vector at time step $t$.

$$\sigma(x) = \frac{1}{1+e^{-x}} \tag{6}$$

Where $\sigma(x)$ is the sigmoid activation function.

**3.3 Transformer Component**

The Transformer model relies on attention mechanisms. It was initially developed for natural language processing tasks such as machine translation and language modelling. The core principle of the Transformer is to capture global dependencies in sequential data through self-attention mechanisms, without the need for RNNs or CNNs(Koohfar et al., 2023). The Transformer model has been widely applied to sequence-to-sequence tasks, demonstrating outstanding performance and scalability.

In renewable energy demand forecasting, the Transformer component, building on the features extracted by the LSTM, is critical in the final prediction process. By leveraging the self-attention mechanism, the Transformer effectively processes global relationships in the sequential data and further optimizes feature representations(Al-Ali et al., 2023). This capability enables the model to better comprehend the significant information within the sequence data and produce accurate predictions. Compared to traditional recurrent neural networks, the Transformer's advantage lies in parallel computation, making it more efficient in handling long sequence data. Additionally, the Transformer model can avoid issues such as vanishing or exploding gradients, resulting in a more stable and reliable training process.

Within the overall model, the Transformer receives the feature representations from the LSTM component. It utilizes the self-attention mechanism to integrate these features in a weighted manner, ultimately generating the energy demand prediction results. The self-attention mechanism in the Transformer model dynamically assigns importance weights to different positions during feature integration. For each input feature, the Transformer computes similarity scores with all other features and utilizes these scores as weights to perform a weighted sum of all features, generating a new weighted feature representation. This self-attention mechanism allows the model to dynamically focus on important features across the entire sequence, without being limited by a fixed window size or sliding window. In this way, the Transformer model effectively captures global dependencies in the sequence data and incorporates these relationships into the final energy demand predictions. Figure 3 provides a detailed illustration of its internal operational mechanisms.

Due to the capabilities of the Transformer model, it effectively utilizes global information in the sequence data, thereby enhancing the overall model's predictive performance and generalization ability. Consequently, the Transformer component is crucial in the LTPNet model, providing essential support for accurate predictions and sustainable energy management.

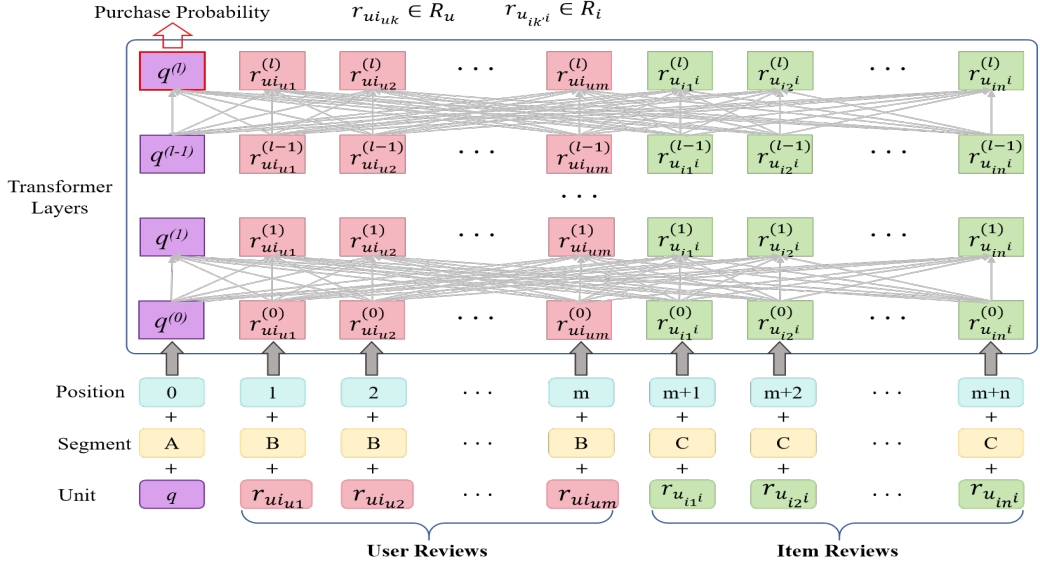

Figure 3. Flowchart of the Transformer model. The Transformer model consists of multiple layers of encoders and decoders, each of which uses a self-attention mechanism to capture the complex relationship between input data. The user comments and product reviews in the figure are encoded into sequence inputs, which are processed by position encoding, fragment encoding, and unit encoding before entering the multi-layer Transformer structure for feature extraction.

The main formula of Transformer model is as follows:

$$\text{Encoder Output} = \text{MultiHead Attention}(Query, Key, Value) \tag{7}$$

Where $Query$, $Key$, and $Value$ are input sequences, and $MultiHead\ Attention$ is the multi-head attention mechanism.

$$\text{Positional Encoding} = [\sin(pos/10000^{2i/d})\cos(pos/10000^{2i/d})] \tag{8}$$

Where pos is the position and d is the dimension of the positional encoding.

$$\text{Decoder Input} = \text{MultiHead Attention}(Query, Key, Value) \tag{9}$$

Where $Query$, $Key$, and $Value$ are output sequences from the encoder.

$$\text{Feed Forward}(X) = \text{ReLU}(X \cdot W_1 + b_1) \cdot W_2 + b_2 \tag{10}$$

Where $X$ is the input, $W_1$, $b_1$, $W_2$, and $b_2$ are learnable parameters.

$$\text{Layer Norm}(X) = \frac{X - \mu}{\sigma}, \text{where } \mu = \text{mean}(X), \sigma = \text{std}(X) \tag{11}$$

Where $X$ is the input tensor, and $\mu$ and $\sigma$ are the mean and standard deviation, respectively.

### 3.4 PSO Optimization

Particle Swarm Optimization (PSO) is an optimization algorithm inspired by the collaborative and competitive behavior observed in biological swarms, such as flocks of birds or schools of fish. It aims to identify the optimal solution by simulating the interactions among individuals within the swarm(Vinothkumar et al., 2023). In the optimization of LSTM and Transformer parameters, PSO is utilized to explore the most effective combinations of hyperparameters, thereby improving the model's predictive performance.

In the context of renewable energy demand forecasting, PSO is extensively used to refine the parameters of both deep learning and traditional machine learning models. Its key advantage lies in its ability to efficiently navigate through a broad parameter space without relying on explicit gradient information.(Eldeghady et al., 2023). The PSO algorithm also has strong global search capabilities and good convergence properties, enabling it to find near-optimal parameter combinations.

In the context of the overall model, the PSO component plays a crucial role. Initially, the PSO algorithm is utilized to refine the hyperparameters of the LSTM model, such as the number of hidden units and the learning rate. This optimization aims to enhance the model's compatibility with the data and boost its predictive capabilities. During the PSO optimization process, the collaboration among individuals and the swarm helps the model converge more quickly to a near-optimal solution, thus enhancing predictive performance. Moreover, the PSO optimization method is also utilized to fine-tune the hyperparameters of the Transformer model, aiming to enhance its performance further. This process involves adjusting parameters such as the number of layers, attention heads, and hidden units in the Transformer architecture. These adjustments aim to improve the model's ability to capture global dependencies within the sequential data.

The PSO optimization component is vital in the LTPNet model. It searches the parameter space for optimal hyperparameter combinations, thereby enhancing the model's performance and generalizability. Figure 4 clearly illustrates the flow structure of the PSO algorithm.

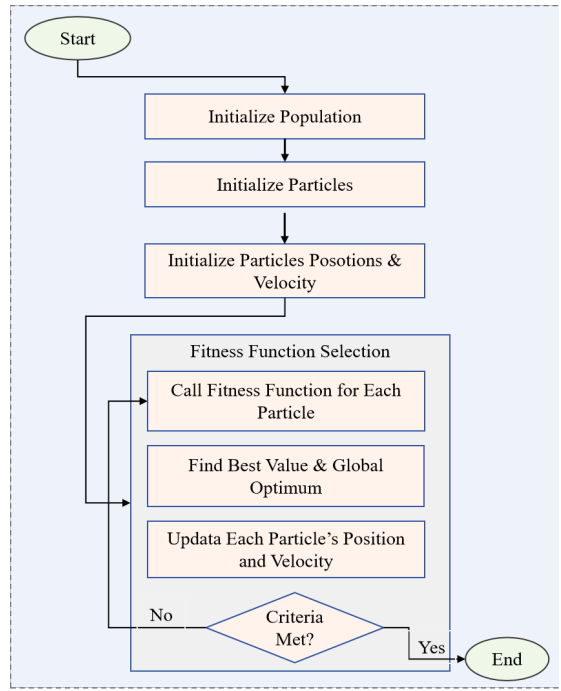

**Figure 4. Flow chart of the PSO Model. Initialize the population and the initial position and velocity of each particle. Then, evaluate the performance of each particle according to the fitness function to find the current optimal value and the global optimal solution. In each iteration, update the position and velocity of each particle. By repeatedly calling the fitness function and adjusting the distribution of the particle population according to the new position and velocity, the stopping criterion is met.**

The main formula of PSO model is as follows:

$$x_i(0) = x_i^{min} + rand() \cdot (x_i^{max} - x_i^{min}) \quad (12)$$

Where $x_i(0)$ is the initial position of particle $i$, $x_i^{min}$ and $x_i^{max}$ are the minimum and maximum bounds for the position, and $rand()$ is a random number between 0 and 1.

$$v_i(0) = 0 \quad (13)$$

Where $v_i(0)$ is the initial velocity of particle $i$.

$$v_i(t+1) = \omega \cdot v_i(t) + c_1 \cdot rand() \cdot (p_i - x_i(t)) + c_2 \cdot rand() \cdot (g - x_i(t)) \quad (14)$$

Where $v_i(t+1)$ is the velocity of particle $i$ at time $t+1$, $\omega$ is the inertia weight, $c_1$ and $c_2$ are cognitive and social coefficients, $p_i$ is the best previous position of particle $i$, $g$ is the global best position among all particles, and $rand()$ is a random number between 0 and 1.

$$x_i(t+1) = x_i(t) + v_i(t+1) \tag{15}$$

Where, $x_i(t+1)$ is the position of particle $i$ at time $t+1$.

$$p_i = \begin{cases} x_i(t+1) & if\ f(x_i(t+1)) < f(p_i) \\ p_i & otherwise \end{cases} \tag{16}$$

Where $f()$ is the objective function.

$$g = \underset{x}{\mathrm{argmin}}\, f(x_i(t+1)), \forall i \tag{17}$$

Where $g$ is updated if any particle finds a position with a lower objective function value.

$$f(x) = \sum_{i=1}^{n} x_i^2 \tag{18}$$

Where $f(x)$ represents an example objective function to be minimized by the PSO algorithm, and $n$ is the dimension of the search space.

## 4 Experiment

### 4.1 Dataset

Among the datasets used in this study are the Global Wind Atlas data, solar resource datasets, renewable energy production datasets, and electricity market transaction datasets. These datasets, sourced from a wide range of regions and time periods, provide rich information and experimental data for the research.

The Global Wind Atlas Data, developed collaboratively by the World Bank and the International Renewable Energy Agency (IRENA) among others, is a high-quality dataset of wind energy resources that covers the entire globe. It provides detailed information on wind speed, wind direction, and wind power density with high spatial and temporal resolution(Davis et al., 2023). Utilizing advanced numerical weather prediction models and meteorological observation data, the dataset ensures high accuracy and reliability through meticulous calculations and analyses. Data on wind speeds at various altitudes, wind direction distribution, and wind power density offers a scientific basis for policymakers, energy planners, investors, and developers. It supports the assessment, forecasting, and development of wind energy resources, reduces investment risks, and provides critical evidence for the experiments and forecasts described in this paper.

The solar resource dataset contains solar radiation data from different geographical regions and time periods(Lagili et al., 2023). The data comes from multiple solar resource monitoring stations and meteorological stations, and the data features include information such as solar radiation intensity, solar altitude angle, and solar azimuth angle. These data are used to evaluate and predict the availability of solar resources. In terms of data diversity, the dataset covers solar energy data under different climate conditions, latitudes and seasons, which guarantees the generalization ability of the model under various weather and environmental conditions.

The renewable energy production dataset includes power generation data of renewable energy such as wind and solar energy in different regions. The data comes from the real-time monitoring system of the power company and the operation data of renewable energy power plants(Alcántara et al., 2023). It has a high temporal resolution and can reflect the renewable energy production in different geographical regions and time periods. Data collection methods include real-time monitoring and regular reporting, which ensure the accuracy and reliability of the data. The diversity of this dataset covers different types of renewable energy, geographical regions and time spans, providing full support for the performance of the model under various energy production conditions.

The power market transaction dataset provides information such as prices, supply and demand relationships and transaction volumes in the power market, covering data from different regions and time periods(Zhang & Li, 2023). The data comes from the public data of power market operators and regulators, and the collection and release of data are highly standardized and transparent. These datasets help analyze and predict the operation of the power market, evaluate the impact of supply and demand changes on market prices and transaction volumes, and provide generalization support for the model under different market conditions.

Overall, the datasets used in this study provide comprehensive and high-quality coverage of global wind resources, solar resources, renewable energy production, and electricity market transactions. These datasets offer comprehensive information and serve as a robust foundation for conducting experiments, providing essential assistance for training and evaluating the model.

**4.2 Experimental Setup**

To maintain precision and consistency in the experiment, this article established meticulous configurations for the experimental setup.

**Step 1**: Data preprocessing

- Data Cleaning: For data with more than 5% missing values, we use the mean or median to fill in. The three-times standard deviation principle is used to identify and remove outliers to improve the quality and accuracy of the data.

- Data Standardization: The Z-score standardization method is used to convert the data into a standard normal distribution with a mean of 0 and a standard deviation of 1 to eliminate the impact of different data dimensions on model training.

- Data Splitting: The dataset is divided into a training set and a test set with a ratio of 7:3 for model training and performance evaluation.

**Step 2:** Model Training

- Network parameter settings: For the LSTM model, 128 hidden units, a learning rate of 0.001, and a batch size of 64, are defined.. As for the Transformer model, it consists of six layers, eight attention heads, and 256 hidden units, with a learning rate fixed at 0.0001. The PSO optimization algorithm is configured with 50 particles, 100 iterations, an inertia weight of 0.5, and learning factors set to 1.

- Model architecture design:  The model consists of three components: LSTM, Transformer, and PSO. LSTM processes historical energy consumption data to capture temporal dependencies and long-term trends; the Transformer integrates features extracted by LSTM to generate the final energy demand prediction; PSO optimizes the hyperparameters of LSTM and Transformer to enhance the overall model's predictive performance. The design considers the synergy and complementarity of the three components to further improve predictive performance and generalization ability. The algorithm process of the LTPNet model as shown in Table 1.

| **Algorithm 1**: Training Process of LTP-Net model |
|---|
| **Input**: Training dataset:$\{X_i, Y_i\}_{i=1}^{N}$ Learning rate $\eta$, Number of epochs: $E$ |
| **Output**: Trained parameters:$\Theta$ |
| Initialize LSTM parameters: $\Theta_L$; |
| Initialize Transformer parameters: $\Theta_T$; |
| Initialize PSO parameters:$\Theta_P$; |
| **for** $epoch \leftarrow 1$ **to** E **do** |
|     **for** $i \leftarrow 1$ **to** N **do** |
|         Sample batch data:$\{X_a, Y_b\}$; |
|         // Forward Pass |
|         $H_L = LSTM(X_b, \Theta_L)$; $H_T = Transformer(H_L, \Theta_T)$; $H_{pred} = PSO(H_T, \Theta_P)$; |
|         // Loss Calculation |
|         $Loss = MSE(Y_{pred}, Y_b)$; |
|         // Backward Pass |

Update LSTM parameters: $\Theta_L \leftarrow \Theta_L - \eta \cdot \nabla_{\Theta_L} Loss$;
Update Transformer parameters: $\Theta_T \leftarrow \Theta_T - \eta \cdot \nabla_{\Theta_T} Loss$;
Update PSO parameters: $\Theta_P \leftarrow \Theta_P - \eta \cdot \nabla_{\Theta_P} Loss$;
   end
end
**Return** Trained parameters: $\Theta = \{\Theta_L, \Theta_T, \Theta_P\}$;

**Table 1 The algorithm process of the LTPNet model.**

● Model training process: The Stochastic Gradient Descent (SGD) algorithm optimizes the model parameters to minimize the loss function. Our training procedure comprises 100 epochs, each with 1000 iterations, ensuring sufficient learning opportunities and convergence. Additionally, to prevent overfitting, we implement an early stopping strategy that terminates training when the loss on the validation set no longer decreases.

**Step 3:** Model Validation and Tuning

● Cross-validation: The training dataset is divided into five mutually exclusive subsets. Four are used for training and one for validation. This process is repeated five times with different validation sets. This allows for a thorough assessment of the model's performance, offering a comprehensive evaluation.

● Model fine-tuning: Grid Search explores different parameter combinations to select the best-performing set. Throughout the experiments, we explored different combinations of learning rates (0.001, 0.0001), batch sizes (32, 64), and layer numbers (4, 6) to determine the most effective model parameters..

**Step 4:** Ablation Experiment

Three distinct experimental conditions assess the individual impact of each component in the LTP-Net model.

● Removing the LSTM Component: Exclude the LSTM component from the model architecture, with only the Transformer and PSO components retained. The LSTM's hidden units were effectively set to zero, thus eliminating its influence on the overall model. The parameters of the Transformer and PSO models remained unchanged.

● Removing the Transformer Component: Exclude the Transformer component was removed, with only the LSTM and PSO components retained. All parameters of the Transformer model were set to 0, effectively eliminating the influence of the Transformer model. The parameters of the LSTM and PSO models remained unchanged.

● Removing PSO: Exclude the PSO optimization component was removed, with only the LSTM and Transformer components retained. In this scenario, we nullified all parameters associated with the PSO optimization algorithm, effectively negating its impact on the overall process. Meanwhile, the parameters pertaining to the LSTM and Transformer models were left unaltered.

**Step 5:** Comparative Analysis

A series of experiments evaluate the impact of different optimization strategies on the performance of the LTP-Net model:

● Adam vs. PSO: The Adam optimizer, a commonly used gradient-based optimization algorithm, is compared with PSO. Adam is configured with a learning rate of 0.001 and a batch size of 64, using default parameters

● Bayesian Optimization vs. PSO: Bayesian optimization is set with an initial sampling of five iterations and ten subsequent iterations, using default parameters. The PSO parameters remain unchanged.

● Self-AM vs. PSO: The Self-Adaptive Momentum optimization algorithm is configured with an initial momentum parameter of 0.9, a momentum update rate of 0.1, a learning rate of 0.001 and a batch size of 64. The PSO parameters remain unchanged

**Step 6:** Model Evaluation

A comprehensive array of accuracy and efficiency metrics, including model parameters, floating-point operations (FLOPs), inference time, and training time, are employed to gauges the performance of the LTP-Net model. These metrics included Mean Absolute Error (MAE), Mean Absolute Percentage Error (MAPE), Root Mean Square Error (RMSE), and Mean Squared Error (MSE). By comparing the model's predictions to actual values, these metrics offered a detailed assessment of its predictive accuracy and highlighted any disparities.

Efficiency metrics, such as model parameters, floating-point operations (FLOPs), inference time, and training time, were taken into consideration. These metrics assessed the computational complexity and efficiency of the model, reflecting its inference speed and training efficiency. By thoroughly analyzing these evaluation metrics, we gained a comprehensive understanding of the LTP-Net model's performance in terms of predictive accuracy and computational efficiency. This analysis provides valuable insights for further model optimization and improvement.

**4.3 Experimental Result and Analysis**

The results shown in Tables 2 and 3 show that our LTP-Net model demonstrates significant advantages across various datasets, including the Global Wind Atlas Data, Solar Resource Dataset, Renewable Energy Production Dataset, and Electricity Market Transaction Dataset. Taking the MAE metric as an example, on the Global Wind Atlas Data, our model achieves an MAE of only 15.19, while other models exhibit MAE values ranging from 22.21 to 50.27. Similarly, our model consistently displays lower MAE, MAPE, RMSE, and MSE values across all datasets, clearly reflecting its superiority in prediction accuracy.

Further analysis of the table data reveals that our model consistently yields relatively lower values across performance metrics, indicating enhanced accuracy and reliability in energy demand forecasting. Importantly, our model not only excels on wind and solar energy datasets but also performs remarkably well on renewable energy production and electricity market transaction datasets, demonstrating its versatility and robustness across different types of energy datasets.

| Model | Global Wind Atlas Data | | | | Solar Resource Dataset | | | |
|---|---|---|---|---|---|---|---|---|
| | MAE | MAPE | RMSE | MSE | MAE | MAPE | RMSE | MSE |
| Benti(Benti et al., 2023) | 50.27 | 10.71 | 7.57 | 12.47 | 40.94 | 13.78 | 6.78 | 13.96 |
| Kim(Kim & Kim, 2023) | 32.97 | 9.65 | 6.65 | 21.07 | 42.95 | 11.87 | 5.37 | 17.9 |
| Abou Houran(Abou Houran et al., 2023) | 22.58 | 9.47 | 6.75 | 17.14 | 27.26 | 10.82 | 6.73 | 15.49 |
| Mansouri(Mansouri et al., 2023) | 42.35 | 8.95 | 8.2 | 22.77 | 48.64 | 10.66 | 5.88 | 18.57 |
| Nwokolo(Nwokolo et al., 2023) | 22.21 | 8.98 | 7.17 | 29.31 | 44.07 | 9.09 | 7.11 | 28.72 |
| Aksan(Aksan et al., 2023) | 35.68 | 8.85 | 7.48 | 15.31 | 30.26 | 11.17 | 8 | 16.1 |
| Ours | 15.19 | 7.51 | 3.48 | 10.38 | 17.27 | 8.2 | 3.79 | 10.55 |

**Table 2 The comparison of different models in MAE, MAPE, RMSE, MAE indicators comes from Global Wind Atlas Data and Solar Resource Dataset**

| Model | Renewable Energy Production Dataset | | | | Electricity Market Transaction Dataset | | | |
|---|---|---|---|---|---|---|---|---|
| | MAE | MAPE | RMSE | MSE | MAE | MAPE | RMSE | MSE |
| Benti(Benti et al., 2023) | 29.25 | 15.13 | 5.76 | 28.49 | 22.56 | 11.57 | 8.41 | 28.58 |
| Kim(Kim & Kim, 2023) | 22.74 | 10.85 | 6.21 | 24.86 | 47.39 | 11.84 | 5.85 | 18.75 |
| Abou Houran(Abou Houran et al., 2023) | 30.1 | 11.91 | 7.97 | 14.96 | 22.8 | 9.96 | 7.48 | 26.56 |
| Mansouri(Mansouri et al., 2023) | 48.91 | 10.48 | 7.77 | 15.22 | 30.68 | 13.79 | 4.96 | 26.92 |
| Nwokolo(Nwokolo et al., 2023) | 34.57 | 9.67 | 6.03 | 27.52 | 49.33 | 14.25 | 8.34 | 15.5 |
| Aksan(Aksan et al., 2023) | 40.91 | 11.89 | 6.65 | 18.78 | 28.93 | 8.43 | 8.55 | 24.43 |
| Ours | 19.95 | 6.75 | 3.9 | 11.93 | 17.98 | 8.06 | 4.58 | 10.07 |

**Table 3 The comparison of different models in MAE, MAPE, RMSE, MAE indicators comes from Renewable Energy Production Dataset and Electricity Market Transaction Dataset**

To provide a visual comparison of the performance of different models across various metrics, Figure 5 is presented in the form of bar charts. The bar charts vividly depict the superior performance of our LTP-Net model, showcasing lower values across metrics such as MAE, MAPE, RMSE, and MSE compared to other models. This visual representation effectively reinforces the points made in our analysis of experimental results, emphasizing the significant advantage of our model in predicting renewable energy demand for green operations in enterprises compared to existing models.

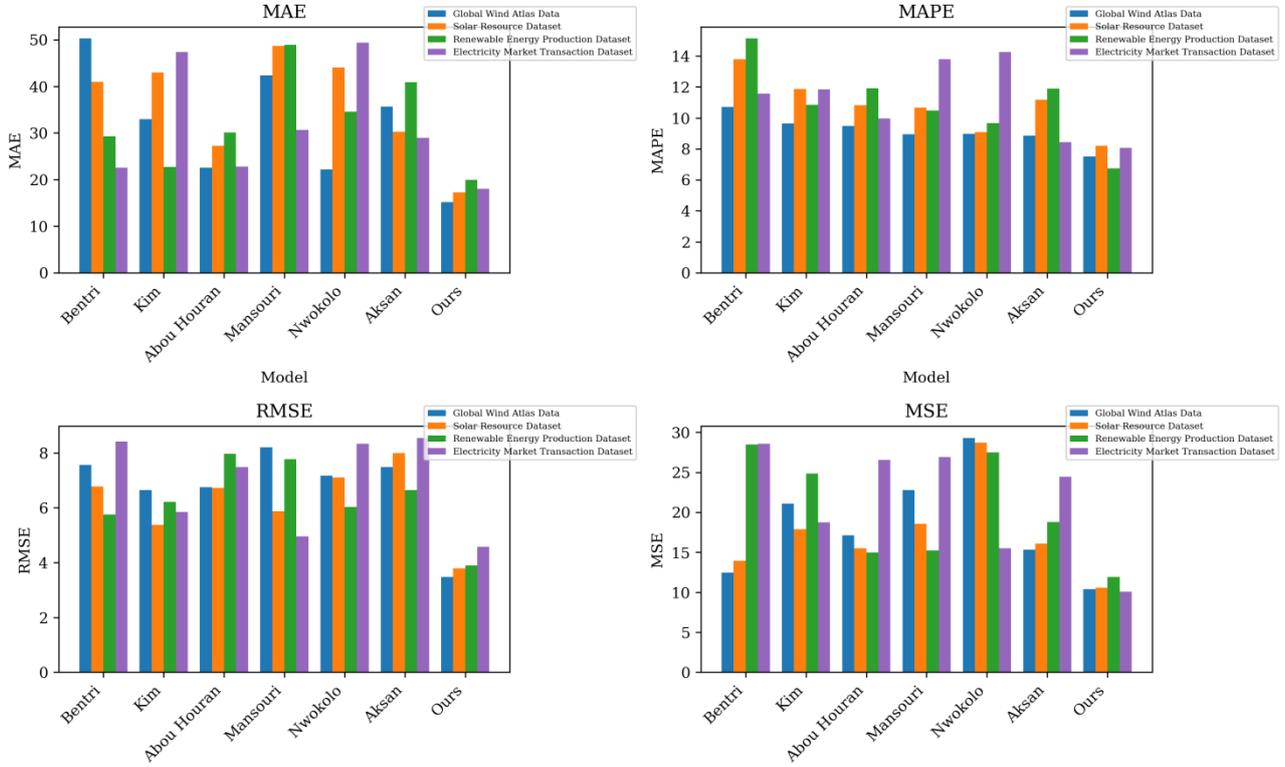

**Figure 5**. **Comparative visualization of the prediction performance of the LTP-Net model and other baseline models on four datasets.**

Additionally, Tables 4 and 5 present an evaluation and comparison of various models on different datasets. Firstly, our LTP-Net model demonstrates significant advantages in metrics such as parameter count, floating-point operations (Flops), inference time, and training time on the Global Wind Atlas dataset. Compared to other models, ours demonstrated lower parameter count and computational load, as well as shorter inference and training times, indicating higher efficiency and speed.

Similarly, our model showed good performance on the Solar Resource Dataset, Renewable Energy Production Dataset, and Electricity Market Transaction Dataset. Although some models had slight advantages on certain datasets, overall, our model maintained stable performance across various metrics, indicating its good generalization and robustness.

| Model | Global Wind Atlas Data | | | | Solar Resource Dataset | | | |
| --- | --- | --- | --- | --- | --- | --- | --- | --- |
|  | Parameters (M) | Flops (G) | Inference Time (ms) | Trainning Time (s) | Parameters (M) | Flops (G) | Inference Time (ms) | Trainning Time (s) |
| Benti(Benti et al., 2023) | 516.22 | 6.31 | 7.56 | 540.85 | 487.27 | 6.18 | 8.15 | 489.03 |
| Kim(Kim & Kim, 2023) | 750.28 | 7.29 | 13.04 | 725.92 | 747.39 | 7.71 | 12.13 | 639.82 |
| Abou Houran(Abou Houran et al., 2023) | 614.11 | 5.26 | 6.38 | 713.19 | 725.95 | 4.97 | 6.55 | 547.93 |
| Mansouri(Mansouri et al., 2023) | 646.63 | 7.67 | 9.64 | 629.31 | 582.94 | 7.04 | 10.53 | 671.79 |

| Model | Nwokolo(Nwokolo et al., 2023) | | | | | | | |
|---|---|---|---|---|---|---|---|---|
| | 413.24 | 5.27 | 7.30 | 441.62 | 445.36 | 4.93 | 7.23 | 445.95 |
| Aksan(Aksan et al., 2023) | 539.59 | 5.48 | 8.20 | 518.66 | 482.16 | 5.29 | 8.33 | 566.44 |
| Ours | 338.72 | 3.55 | 5.33 | 326.40 | 319.94 | 3.63 | 5.59 | 336.57 |

**Table 4 Model efficiency verification and comparison of different indicators of Global Wind Atlas Data and Solar Resource Dataset**

| Model | Renewable Energy Production Dataset | | | | Electricity Market Transaction Dataset | | | |
|---|---|---|---|---|---|---|---|---|
| | Parameters (M) | Flops (G) | Inference Time (ms) | Trainning Time (s) | Parameters (M) | Flops (G) | Inference Time (ms) | Trainning Time (s) |
| Benti(Benti et al., 2023) | 479.35 | 6.19 | 9.17 | 499.40 | 520.27 | 6.39 | 9.71 | 512.15 |
| Kim(Kim & Kim, 2023) | 667.98 | 8.60 | 13.33 | 780.76 | 627.89 | 8.23 | 11.72 | 780.03 |
| Abou Houran(Abou Houran et al., 2023) | 502.33 | 6.05 | 9.96 | 718.74 | 462.00 | 7.35 | 10.04 | 447.37 |
| Mansouri(Mansouri et al., 2023) | 792.71 | 6.96 | 9.77 | 606.13 | 628.85 | 7.09 | 12.90 | 622.81 |
| Nwokolo(Nwokolo et al., 2023) | 467.85 | 5.12 | 7.36 | 423.67 | 447.68 | 5.32 | 8.25 | 478.07 |
| Aksan(Aksan et al., 2023) | 519.57 | 5.15 | 8.39 | 474.27 | 486.20 | 5.75 | 8.11 | 489.54 |
| Ours | 338.04 | 3.85 | 5.37 | 329.81 | 318.40 | 3.66 | 5.63 | 337.38 |

**Table 5 Model efficiency verification and comparison of different indicators of Renewable Energy Production Dataset and Electricity Market Transaction Dataset**

For a concise illustration of the performance of different models across the four datasets, Figure 6 provides a visual representation of the data. It shows that our LTP-Net model falls within a moderate range in terms of parameter count, effectively controlling model size while maintaining adequate complexity. Although slightly higher in inference and training times compared to some models, given our model's significant performance advantages in prediction, this investment of time is reasonable. Particularly, on the Renewable Energy Production and Electricity Market Transaction datasets, the LTP-Net model demonstrated outstanding inference and training efficiency, validating its effectiveness and practicality in real-world applications. Additionally, the efficient performance of LTP-Net in floating-point operations further emphasizes its capability and efficiency in handling large-scale data. Overall, the analysis from Figure 6 clearly indicates that while ensuring excellent performance, LTP-Net also optimizes the utilization of computational resources.

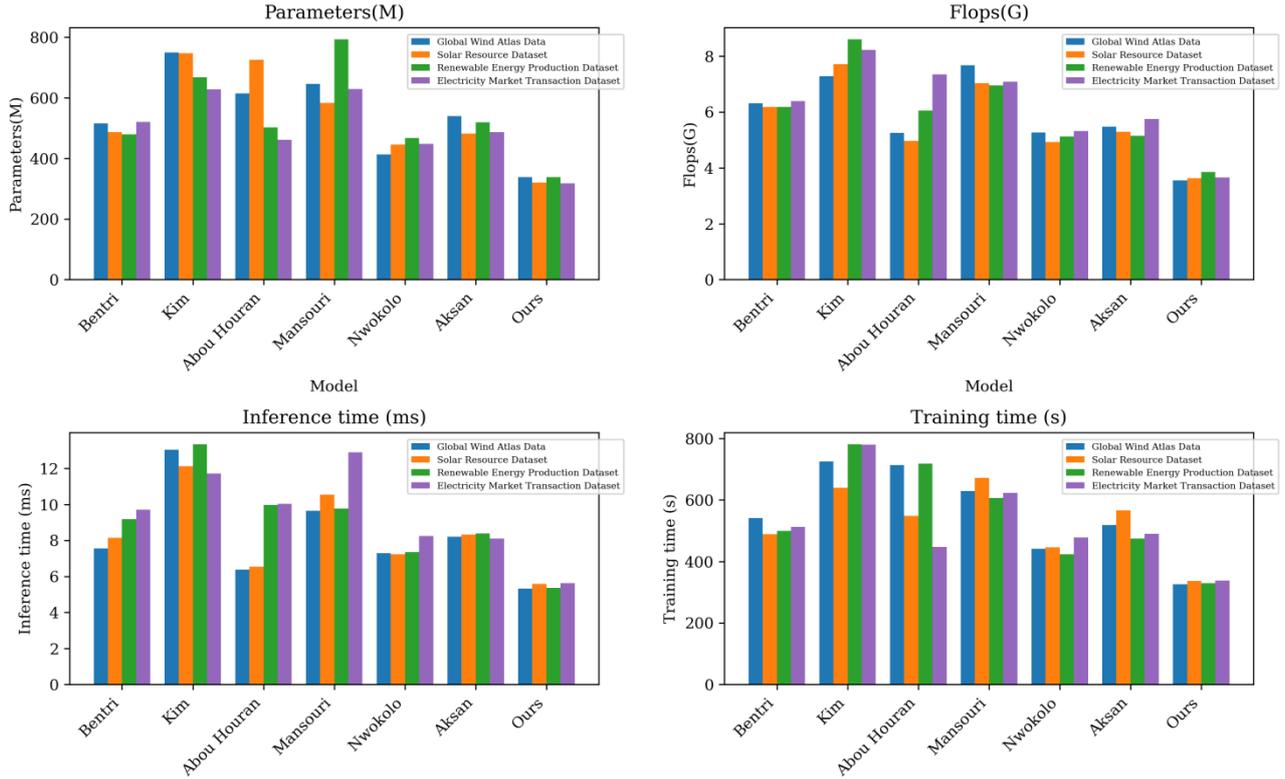

**Figure 6**. **Comparative visualization of the model efficiency of the LTP-Net model and other baseline models on four datasets.**

Through a series of ablation experiments illustrated in Tables 6 and 7, we evaluate the independent contributions of each component within the LTP-Net model. By comparing model performance under different experimental conditions, we can clearly observe the impact of each component on the overall model.

In our experiments, we first removed the LSTM component, retaining only the Transformer and PSO optimization components. The results indicate that the model utilizing only Transformer and PSO performed well across various evaluation metrics, yet it still fell short of the complete model's performance. For instance, in the Global Wind Atlas Data dataset, the MAE for the model with only Transformer and PSO was 40.17, significantly higher than the complete model's MAE of 15.19.

Next, we removed the Transformer component, keeping the LSTM and PSO optimization components. The findings show that the model using only LSTM and PSO performed better on some datasets but did not match the complete model's performance on others. For example, in the Electricity Market Transaction Dataset, the MAPE for the model with only LSTM and PSO was 15.18%, whereas the complete model achieved a MAPE of 8.06%.

Lastly, we removed the PSO optimization component, retaining the LSTM and Transformer components. The results demonstrate that the model with only LSTM and Transformer was adequate in some datasets but overall significantly underperformed compared to the complete model. For example, in the Solar Resource Dataset, the RMSE for the model with only LSTM and Transformer was 8.42, a considerable difference from the complete model's RMSE of 3.79.

The comprehensive comparison of all experimental groups indicates that the complete LTP-Net model excelled across all evaluation metrics, clearly outperforming the models where any single component was removed. This further validates the significance and role of LSTM, Transformer, and PSO within the complete model. Finally, Figure 7 visualizes the content of the tables, clearly displaying the performance differences under various experimental conditions. From this visualization, one can distinctly see the advantages of the complete model compared to other experimental setups.

| Model | Global Wind Atlas Data | Solar Resource Dataset |
|---|---|---|

|                  | MAE   | MAPE  | RMSE | MSE   | MAE   | MAPE  | RMSE | MSE   |
|------------------|-------|-------|------|-------|-------|-------|------|-------|
| Transformer+PSO  | 40.17 | 12.37 | 6.85 | 13.29 | 30.84 | 11.01 | 6.26 | 25.78 |
| LSTM+PSO         | 27.61 | 10.18 | 7.60 | 16.69 | 41.15 | 9.08  | 5.42 | 27.54 |
| LSTM+Transformer | 29.74 | 15.08 | 6.93 | 13.92 | 46.55 | 10.03 | 8.42 | 24.48 |
| ALL(LTP-Net)     | 15.19 | 7.51  | 3.48 | 10.38 | 17.27 | 8.2   | 3.79 | 10.55 |

**Table 6 Ablation experiments on the LTPNet module using Global Wind Atlas Data and Solar Resource Dataset**

| Model | Renewable Energy Production Dataset | | | | Electricity Market Transaction Dataset | | | |
|-------|------|------|------|------|------|------|------|------|
|       | MAE  | MAPE | RMSE | MSE  | MAE  | MAPE | RMSE | MSE  |
| Transformer+PSO  | 27.85 | 12.77 | 5.42 | 20.26 | 45.47 | 11.36 | 4.62 | 12.62 |
| LSTM+PSO         | 39.04 | 9.11  | 5.99 | 21.38 | 46.62 | 15.18 | 5.41 | 28.85 |
| LSTM+Transformer | 21.00 | 8.55  | 8.08 | 14.51 | 28.13 | 15.22 | 7.85 | 17.58 |
| ALL(LTP-Net)     | 19.95 | 6.75  | 3.9  | 11.93 | 17.98 | 8.06  | 4.58 | 10.07 |

**Table 7 Ablation experiments on the LTPNet module using Renewable Energy Production Dataset and Electricity Market Transaction Dataset**

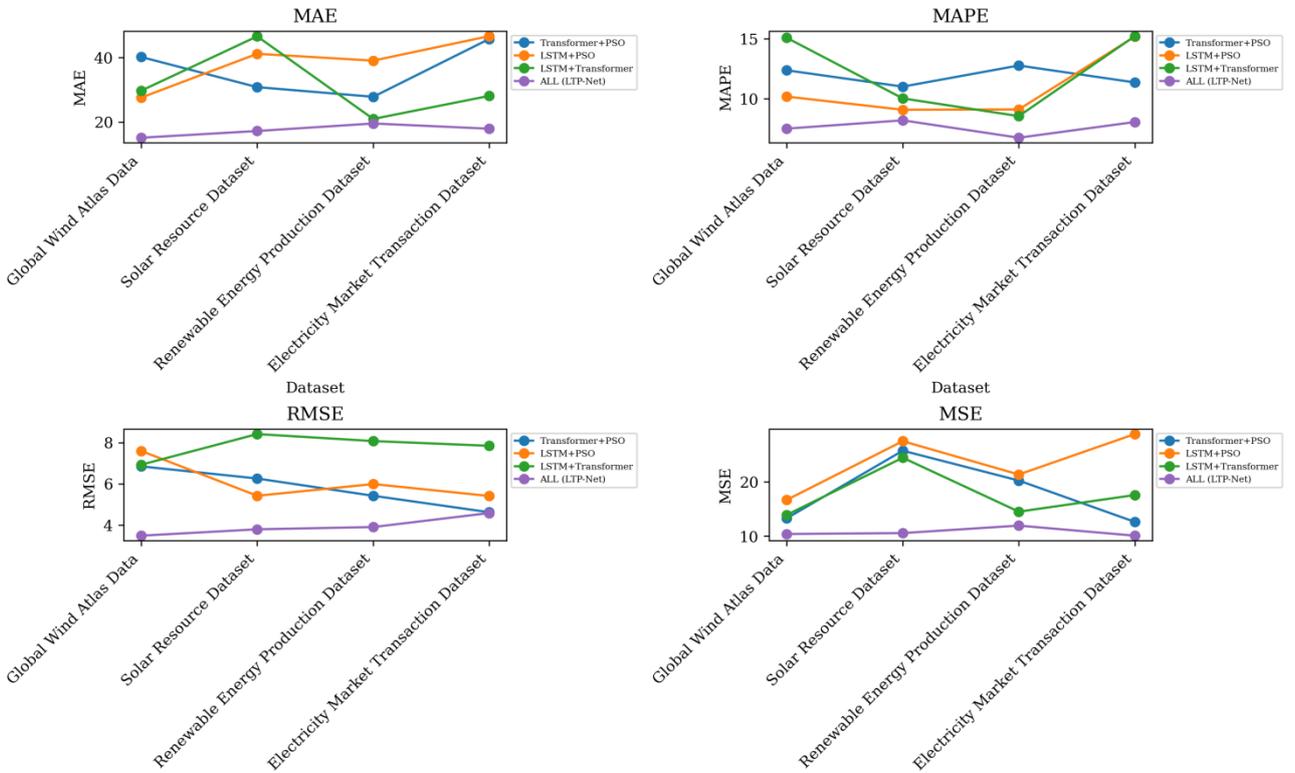

**Figure 7. Ablation experiments on the LTP module**

Furthermore, we conducted a series of comparative experiments to assess the performance impact of four optimization strategies, namely Adam, Bayesian, Cross-AM, and PSO, on the LTP-Net model, as shown in Table 8 and Table 9.

Comparing various metrics, we observe distinct differences between different optimization strategies. Specifically, the PSO optimizer outperforms the Adam optimizer in all metrics, as seen in its superior performance in terms of parameter count, Flops, inference time, and training time on the Global Wind Atlas Data dataset. Further comparing the Bayesian optimization algorithm with the PSO optimization algorithm, we observed that the PSO optimization algorithm performs better across all metrics. For instance, using the Solar Resource Dataset, the PSO optimization algorithm demonstrates significant advantages over the Bayesian optimization algorithm in terms of parameter count, Flops, inference time, and training time. Moreover, compared to the Cross-AM optimization algorithm, the PSO optimization algorithm also shows better performance. For instance, using the Renewable Energy Production Dataset, the PSO optimization algorithm significantly outperforms the Cross-AM optimization algorithm in terms of parameter count, Flops, inference time, and training time.

| Model | Global Wind Atlas Data | | | | Solar Resource Dataset | | | |
| --- | --- | --- | --- | --- | --- | --- | --- | --- |
| | Parameters (M) | Flops (G) | Inference Time (ms) | Trainning Time (s) | Parameters (M) | Flops (G) | Inference Time (ms) | Trainning Time (s) |
| Adam | 375.29 | 269.78 | 254.35 | 308.59 | 365.34 | 383.17 | 207.17 | 407.93 |
| Bayesian | 380.41 | 305.14 | 267.3 | 279.59 | 281.74 | 382.89 | 397.94 | 356.65 |
| Cross-AM | 337.98 | 368.08 | 253.81 | 311.15 | 351.71 | 311.75 | 274.53 | 366.33 |
| PSO | 211.29 | 174.03 | 204.28 | 227.74 | 178.74 | 186.58 | 190.43 | 120.49 |

**Table 8** Ablation experiments on the PSO module using Global Wind Atlas Data and Solar Resource Dataset

| Model | Renewable Energy Production Dataset | | | | Electricity Market Transaction Dataset | | | |
| --- | --- | --- | --- | --- | --- | --- | --- | --- |
| | Parameters (M) | Flops (G) | Inference Time (ms) | Trainning Time (s) | Parameters (M) | Flops (G) | Inference Time (ms) | Trainning Time (s) |
| Adam | 376.11 | 305.1 | 294.6 | 387.9 | 280.82 | 245.1 | 341.76 | 384.12 |
| Bayesian | 372.75 | 267.43 | 246.94 | 284.29 | 381.27 | 293.85 | 220.24 | 395.58 |
| Cross-AM | 305.46 | 325.18 | 233.82 | 286.51 | 350.29 | 288.97 | 385.13 | 403.29 |
| PSO | 155.28 | 136.43 | 213.23 | 191.83 | 206.36 | 222.06 | 210.7 | 187.02 |

**Table 9** Ablation experiments on the PSO module using Renewable Energy Production Dataset and Electricity Market Transaction Dataset

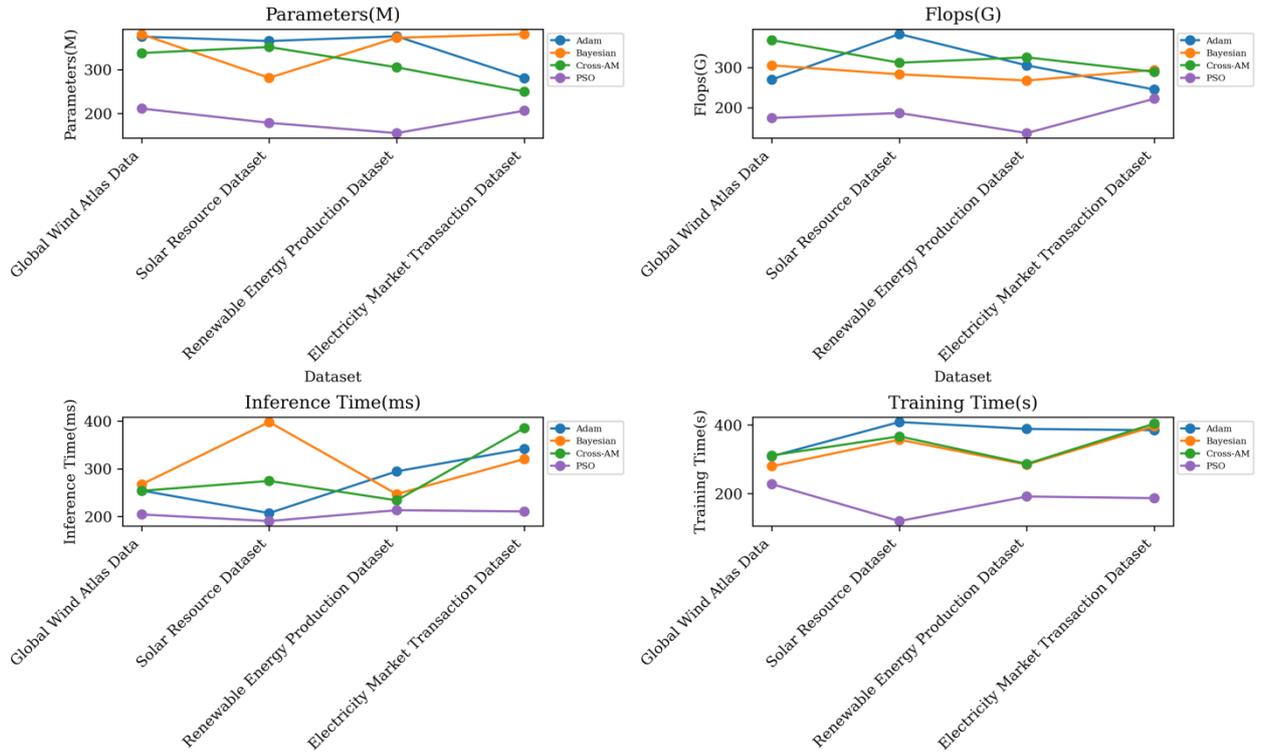

**Figure 8**. Ablation experiments on the PSO module

Figure 8 visualizes the table content, demonstrating the performance differences among various optimization strategies. It is evident from the figure that the performance of each model varies across different datasets, reflecting the adaptability and efficiency of each model towards specific data types. Regarding parameter count, the Mansouri model has the highest parameter count across all datasets, whereas ours has relatively fewer parameters, indicating that these models prioritize efficiency and generalization while maintaining sufficient model complexity. In terms of Flops, the floating-point operation requirements of different models are similar, suggesting that most models achieve some level of optimization in computational complexity. In our analysis of inference time and training time, our model exhibits a significant advantage, particularly in inference time, which is notably lower than other models across all datasets. This not only boosts the practicality of our model but also enhances its feasibility in real-world applications. In essence, the results presented in Figure 8 illustrate the superior and balanced performance of our LTP-Net model across various performance metrics.

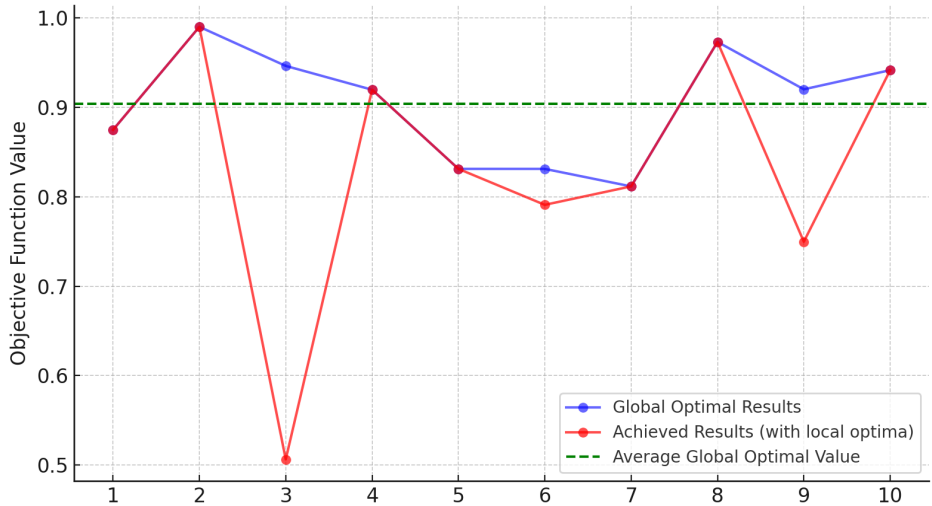

**Figure 9**. **Distribution of optimization results of PSO algorithm under multiple independent runs.**

Figure 9 shows the optimization results of the PSO algorithm under 10 independent runs. The horizontal axis represents the number of each independent experiment, and the vertical axis represents the objective function value obtained from each experiment. The blue curve represents the global optimal solution achieved under ideal conditions, and the red curve represents the results obtained in actual experiments (including some local optimal solutions). The green dashed line shows the average level of the global optimum. Through comparison, it can be seen that the PSO algorithm can find results close to the global optimal solution in most experiments, but in a few cases it will fall into the local optimal solution. This analysis shows that in order to avoid local optimal solutions, the parameter settings of the PSO algorithm need to be further optimized and adjusted.

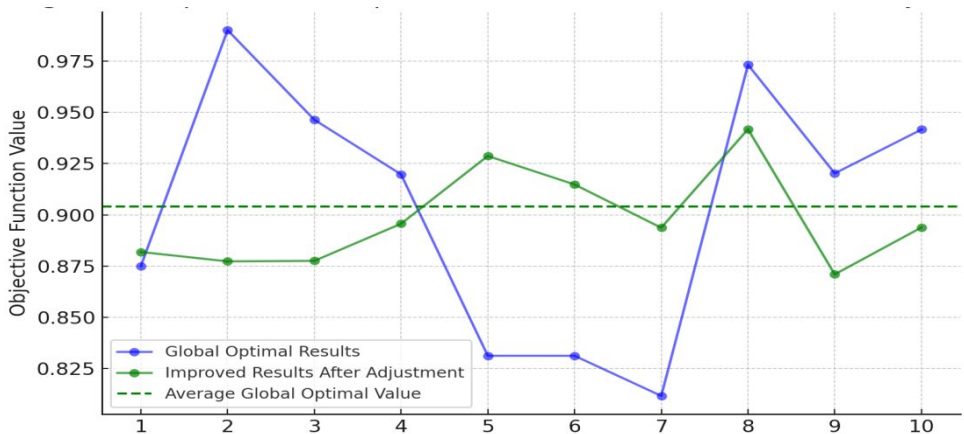

**Figure 10**. **PSO optimization results show the consensus on the global optimal solution.**

Figure 10 shows the optimization results in multiple independent experiments after dynamically adjusting the inertia weight and learning factor of the PSO algorithm. The blue curve represents the theoretical global optimal solution, and the green curve represents the actual results after parameter adjustment. The adjusted results (green curve) show significant improvement. The objective function value in the experiment is closer to the global optimal solution, with less volatility, and the overall performance is more stable. Compared with the situation without parameter adjustment, dynamically adjusting parameters makes the PSO algorithm more consistently approach the global optimal solution in different runs, reducing the possibility of falling into a local optimal solution. This shows that the parameter adjustment strategy adopted effectively enhances the global search capability of the PSO algorithm and improves its optimization performance and reliability under various conditions.

## 5 Conclusion and Discussion

Traditional energy demand forecasting methods are often faced with the problems of complex data processing and low prediction accuracy, which cannot meet the demand of enterprises for renewable energy. In this paper, we propose a method to integrate deep learning technology and environmental decision support system to improve the prediction performance by adopting advanced deep learning techniques such as LSTM and Transformer and combining the PSO algorithm. Experimental results show that our model has advantages under various indicators, which verifies its effectiveness and reliability in renewable energy demand prediction.

Although the LTP-Net model proposed in this paper has achieved significant results in experiments, it has certain limitations. Firstly, the adaptability of the model to specific data types and environmental conditions requires further enhancement, especially its robustness in abnormal situations and extreme weather events. Furthermore, the model's computational efficiency, especially concerning inference and training time, requires further optimization, particularly when handling large-scale datasets.

Future research will be devoted to optimizing the LTPNet model structure and exploring new deep learning models such as Figure neural network to enhance the processing power of complex time series data. At the same time, more advanced optimization algorithms, such as genetic algorithms and Bayesian optimization, will be introduced to further improve the prediction performance and computational efficiency of the model. In addition, the adaptability and robustness of the model under different climatic conditions and regional environments will be verified to ensure its stability and effectiveness in practical applications. The multi-field application prospect of the model is also worth exploring, such as power load forecasting and energy management. Interdisciplinary collaboration will integrate the latest environmental data analysis methods to provide more comprehensive support for energy forecasting.

## Author Contributions

Te Li and Mengze Zhang were responsible for the conceptual design and leadership of the entire project. He led the development of research methods and the design of key experiments. Te Li primarily handled data collection and analysis. Mengze Zhang participated in the processing of experimental data and the statistical validation of results. Yan Zhou was in charge of literature review and theoretical analysis. He also contributed to the writing and revision of the research report, particularly the methods and discussion sections.

## Data availability statement

The data and materials used in this study are not currently available for public access. Interested parties may request access to the data by contacting the corresponding author.

## Conflict of Interests Statement

The authors declare that the research was conducted in the absence of any commercial or financial relationships that could be construed as a potential conflict of interest.

## Funding

No.

## Consent for publication

All authors of this manuscript have provided their consent for the publication of this research.